# Mouse Movement and Probabilistic Graphical Models based E-learning Activity Recognition Improvement Possibilistic Model

Anis Elbahi [#1], Mohamed Nazih Omri[#2], Mohamed Ali Mahjoub [*3] and Kamel Garrouch [#4]

# Research Unit MARS,
Department of computer sciences
Faculty of sciences of Monastir, Monastir, Tunisia.
[1]Elbahi.anis@gmail.com
[2]MohamedNazih.Omri@fsm.rnu.tn
[4]kamelg_2001@yahoo.fr
* Research Unit SAGE,
National Engineering School of Sousse, Sousse, Tunisia.
[3]Medali.mahjoub@ipeim.rnu.tn

*Abstract—* **Automatically recognizing the e-learning activities is an important task for improving the online learning process. Probabilistic graphical models such as Hidden Markov Models and Conditional Random Fields have been successfully used in order to identify a web user activity. For such models, the sequences of observation are crucial for training and inference processes. Despite the efficiency of these probabilistic graphical models in segmenting and labeling stochastic sequences, their performance is adversely affected by the imperfect quality of data used for the construction of sequences of observation.**

**In this paper, a formalism of the possibilistic theory will be used in order to propose a new approach for observation sequences preparation. The eminent contribution of our approach is to evaluate the effect of possibilistic reasoning -during the generation of observation sequences- on the effectiveness of Hidden Markov Models and Conditional Random Fields models. Using a dataset containing 51 real manipulations related to three types of learners' tasks, the preliminary experiments demonstrate that the sequences of observation obtained based on possibilistic reasoning significantly improve the performance of Hidden Marvov Models and Conditional Random Fields models in the automatic recognition of the e-learning activities.**

1. INTRODUCTION

Human activity recognition is an important area of computer vision research [1]. Its applications include a variety of systems that involve interactions between persons and electronic devices such as Human Computer Interfaces. Indeed, identifying web users' activities and preferences especially in a web-based educational environment is crucial for monitoring and interpreting their navigational behavior. Today, online learning has become an important part of society. Therefore, the analysis of learners' navigational behavior plays an important role in the improvement of the learning process. This analysis can be done using, Probabilistic Graphical Models (PGM) which have been successfully used in this field as well as for pattern classification tasks [2, 3]. Two of the most important PGM are the Hidden Markov Models (HMM) and Conditional Random Fields (CRF).

Thanks to their flexibility and elasticity in spatio-temporal pattern recognition tasks and despite all adjustments which have focused on the structure and used algorithms for HMM [12-14] and CRF models [15-18], very few studies have considered the "imperfect" quality of data used by these probabilistic models, specifically in the preparation of sequences of observations used for training and inference. In fact, the data used during training and inference processes are the sources of uncertainty for model predictions [19]. Moreover, the probability distributions on which are based probabilistic models may be affected uncertainty and inconsistency due to the effects of data randomness.

HMM and CRF have been developed to be trained (parameters adjustment) in order to label stochastic temporal sequences. In order to ensure these tasks, the sequences of observations are extremely essential. To generate an observation sequence, let $M=\{x_1, x_2, ..., x_N\}$ be the alphabet of N discrete symbols. Each symbol ($x_i$) is intended to describe "perfectly" at a given time, one state of a real world process. During the generation of each sequence of observations, only one symbol of M will be chosen in order to represent the real state of the dynamic process. Commonly, the choice of this symbol is based on "imperfect" information (knowledge). To better understand this, the following is a simple example describing how to build a sequence of observation based on real observations.

Using the alphabet M={'walking', 'jogging', 'running'} we can create a sequence of observations concerning the sporting activity by observing an athlete in a race. Therefore, $X_1=\{walking_1, jogging_2, jogging_3, …, running_t, …, walking_T\}$ is the observing sequence describing the activity of athlete 1. Indeed, for preparing the sequence $X_1$, the observer (human or machine) noted – at the first time step - that the athlete is in the activity walking=true, jogging=false and running=false. The same observer noted – at the second time step – that the same athlete is in the activity walking=false, jogging=true and running=false.

In reality these activities (walking, jogging, running) are very similar, which can lead another observer to judge the activity –at the first time step - as jogging instead of walking and –at the second time step- as a walking activity instead of jogging. Consequently, the reasoning using the classical approach (true or false), does not take into account uncertainty in the choice of the suitable symbol to describe "perfectly" the real state of the athlete activity. Based on these facts, we propose as a solution to the problem described above, to use the possibilistic reasoning (instead of boolean reasoning) for observation sequences creation for both HMM and CRF models. In this paper, a new observation sequences creation approach based on possibilistic theory is introduced (proposed) and evaluated.

The rest of this paper is organized as follows: Section 2 presents some related works. Section 3 presents activity recognition based on mouse movement. Section 4 briefly introduces HMM and CRF as probabilistic graphical models for activity recognition. Section 5 explains why HMM and CRF data used are uncertain. The possibilistic theory is briefly introduced in section 6. Section 7 presents the proposed possibilistic approach for observation sequences preparation. Section 8 is devoted to present and discuss experimental results. Finally, section 9 concludes the paper and outlines the future works.

## 2. RELATED WORKS

Analyzing the behavior of web users is an important element for satisfying them, giving help and offering what needed and expected. With the increasing performance of computers and the evolution of online user's requirements, various techniques such those proposed in [24-26] have been used in order to understand how and why a user interacts with a web interface. Various studies have been achieved for modeling and analysis of user's behavior during interaction with an online learning application using two of the most popular probabilistic graphical models: HMM and CRF. Courtemanche et al. [27] used HMM approach to recognize the activity of learner within an e-learning platform based on mouse cursor interactions. In [28], HMM have been used for segmenting and labeling mouse trajectory into behavior acts such as cleaning, stopping and searching. Likewise, Soller [33] proposed a new HMM in order to analyze and to assess sequences of online student interaction in order to detect when and why students are having trouble during the e-learning process.

For more than two decades, CRF have been used for segmenting and labeling temporal sequences [11]. Recently, Tong et al.[29] presented a new CRF model for abnormal human activity recognition. Agarwal et al. [30] showed that the discriminative nature of CRF can be exploited to improve the accuracy of recognizing activities. A novel online multitask learning method for human activity recognition using CRF approach was presented by Sun et al. [31]. In order to compare the two techniques (HMM and CRF), some studies used them simultaneously for user activity recognition and discussed why CRF are generally equal to or better than HMM in terms of classification accuracy

[30-32, 35]. The most closely related works, to our work, are the recent studies of Elbahi el al. [34, 35] in which authors proposed two successful HMM and CRF models for inferring the e-learning user tasks based on mouse movement tracking.

The efficiency of HMM and CRF models is extremely dependent on the quality of observations sequences. Each sequence of observations is a data sequence describing a stochastic process relating to a real world application. In fact, during the creation of each sequence of observation, it is assumed that the observer has a perfect knowledge of the observed process, which is not always true in practice due to the uncertainty covering real world data used to prepare each sequence of observation.

In this paper, we propose a new approach based on the possibility theory in order to address the uncertainty of the data used to build sequences of observations, thereafter, the effect of the proposed approach on the effectiveness of HMM and CRF models in web activity recognition will be evaluated.

### 3. ACTIVITY RECOGNITION BASED ON CURSOR MOVEMENTS TRACKING

The automatic identification of the online activities is an important task to improve the general human machine interaction process. This improvement can be made by giving help in real time to unfamiliar users, evaluating online user's needs, helping users with disabilities or improving systems security and interfaces usability.

To perform a given activity, without being aware, users interact with an interface by moving the mouse cursor across web interfaces. Each interface can be defined as a set of significant regions commonly known as Areas Of Interest (AOI) or Items Of Interest (IOI) which can be manually specified [27] or automatically discovered [36]. An AOI may be a link, a particular region in the interface, a picture or other interface element that can draw the users' attention.

It is evident that the mouse device is the most commonly used tool -by computer users- during human computer interaction process. Likewise, the activity of mouse cursor can be easily captured, recorded and analyzed in order to prove high quality clues of a spontaneous, precise, direct and unbiased trace of user behavior. In fact, the analysis of this trace can lead to about the task performed by the user. It also shows that user's attention is drawn to some areas of interest more than others during interaction process [34, 37]. Consequently, for each area corresponds an attraction power called "target gravity" [38] which attracts the cursor in this interface item. A higher attraction power tells us about the higher relevance of an area of interest. Pool and Fitts [39, 37] found that the importance of a given AOI can be defined based on the total fixation number recorded in this area. Therefore, a higher fixation number involves a higher attractive power. It should also be noted that during each activity, using a pointing device a user can frequently fix some areas and ignore others. Although the order of fixing areas is quite random, it reflects the strategy of the user during task realization.

For HMM and CRF models, the sequences of observations are crucial for parameter estimation and inference process. Classically, sequences of observations are generated based on Boolean reasoning. For example the observation sequences used by models developed by Elbahi et al. [34, 35, 40] for user task recognition are based on total belonging of a cursor fixation to an AOI. So, if a fixation is inside an area of interest then this area will be instantiated in the sequence of observations, else the fixation will be totally ignored and not instantiated in the observations sequence. In this paper we propose to use the possibilistic approach for dealing with fixations which are outside areas of interest but near to them, to discover if their consideration in the generation of observation sequences will improve the recognition rate of HMM and CRF models. We call our new models based on possibilistic sequences of observation, PHMM and PCRF respectively for Possibilistic Hidden Markov Model and Possibilistic Conditional Random Fields model.

The next section briefly describes the HMM and CRF approaches and the uncertainty covering the used data during the generation of the observing sequences.

### 4. HMM AND CRF: A BRIEF PRESENTATION

Probabilistic Graphical Models (PGM) are a marriage between graph theory and probability theory [41]. They use a graph-based representation to express the conditional dependence structure between random variables. PGM regroups a variety of models, which can be classified into two major categories: the first one is the Directed Probabilistic Graphical Models (DPGM) and the second category is the Undirected Probabilistic Graphical Models (UPGM). For each PGM the three fundamental cornerstones are representation, inference, and parameters estimation.

*4.1 Hidden Markov Models:*

HMM have been introduced based on the mathematical foundations of Andrei Andreevich Markov [4, 5] as a probabilistic mathematical framework for modeling and labeling stochastic temporal sequences [6-8]. It should be noted that the first successful implementation of the A.A Markov findings has been made by Baum et al. who dealt with statistical calculations for HMM parameters estimation [9]. Sometime later, Baum et al. developed the "Baum-Welch algorithm" for HMM training which is an Expectation Maximization algorithm based on the forward-backward procedure well described in [10]. In the left side of figure 1 a Hidden Markov Model is represented as a DGPM. In fact, a HMM is described as a generative stochastic process where the evolution is managed by states. The series of states in a Markov chain is not directly observable (hidden states) but can be observed only through another process that produces the sequence of possible observations [42]. To be practicable in real world applications, HMM approach assumes that each label ($y_t$) depends only on its previous label ($y_{t-1}$) and each observation ($x_t$) depends on the current label ($y_t$) as shown graphically in figure 1. Formally, a HMM defined over a set of N hidden states and an alphabet of discrete symbols can be specified by $\lambda = (A, B, \Pi)$ with:

• A={$a_{ij}$}: Matrix of transitions probabilities representing the probability of going from the state i to the next state j.

• B={$b_j(k)$}: Matrix of observations emission probabilities representing the probability that observation k was generated by the state j.

• $\Pi$={$\Pi_i$}: Initial probability distribution vector over initial states.

Rabiner et al. [6] explained the three key problems of interest that must be solved by a HMM to be useful in real world application given a model $\lambda = (A, B, \Pi)$ and an observation sequence X={$x_1, x_2, \ldots, x_T$}. The first problem is the parameters estimation. Indeed, parameters A, B and $\Pi$ are adjusted by maximizing the joint probability P(Y, X) in the training dataset, using Baum-Welch algorithm. P(Y, X) can be represented by the next equation (1) as follows:

$$P(Y, X) = \prod_{t=1}^{T} P(x_t|y_t)P(y_t|y_{t-1}) \qquad (1)$$

After the training process, the trained model can be used to resolve the other two problems:

- To find the sequence of labels that best explains a new observation sequence using Viterbi algorithm.
- Or to calculate the probability of a given observation sequence using forward-backward algorithm.
For further details on HMM, see document of Rabiner [6].

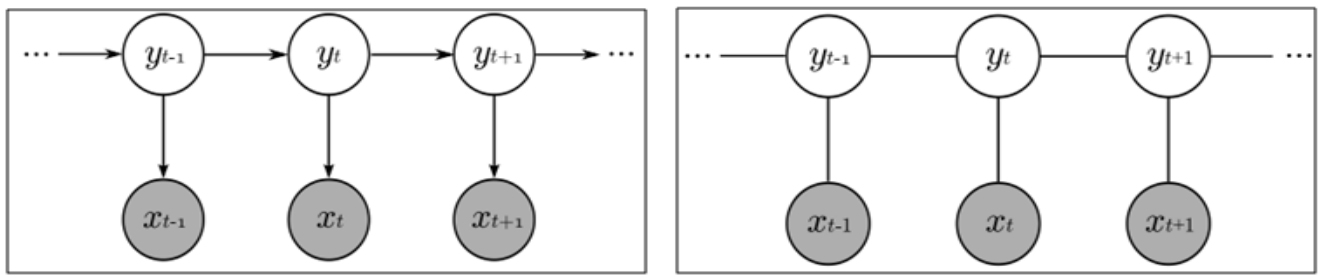

**Fig. 1. Comparison between HMM (left) and CRF (right) graphical representations. Shaded nodes represent observed variables and white nodes represent hidden variables at each time step.**

*4.2 Conditional Random Fields*

CRF was initially introduced by Lafferty et al. [11] as a framework for segmenting and labeling sequence data similarly to HMM but using a different principle. The success of these probabilistic models is a logical consequence of their mere implementation and the robustness of their algorithms used in training and inference processes. As shown on the right side of figure 1, a CRF involves hidden and observable variables at each time step, but they have a different graphical representation than HMM because the edges between nodes are not oriented, making the CRF an UPGM. Unlike HMM, with CRF we try to maximize a conditional probability P(Y|X) instead of joint probability

P(Y, X). By relieving the assumptions of HMM, it becomes possible to represent much more knowledge in the CRF model using feature functions that are difficult to represent in a HMM [30].

The conditional probability distribution P(Y|X) can be described by the equation (2) as follows:

$$P(Y|X) = \frac{1}{Z(X)} exp\left(\sum_{t=1}^{T}\sum_{k=1}^{N} \lambda_k f_k(y_{t-1}, y_t, X) + \sum_{t=1}^{T}\sum_{k=1}^{N} \mu_k g_k(y_t, X)\right) \quad (2)$$

where:
- Z(X) is a normalization factor used to ensure that outcome of P(Y|X) is a probability,
- T is the length of the sequence X,
- N is the number of features functions,
- $\lambda_k$ and $\mu_k$ are weights of each feature function,
- $f_k(y_{t-1}, y_t, X)$ and $g_k(y_t, X)$ are features functions that return a real value. In the simple case, feature functions return 1 or 0.

Despite the different nature of HMM and CRF, both models are very similar because $\lambda_k f_k(y_{t-1}, y_t, X)$ are similar to transition probability $P(y_t|y_{t-1})$ and $\mu_k g_k(y_t, X)$ are analogous to observation probability emission $P(x_t|y_t)$.

CRF are designed to solve the following two problems: the first one is the estimation of model parameters using an iterative gradient method such as BFGS algorithm and the second is the inference process done by Viterbi algorithm. For CRF parameters estimation, we use a training set D, defined by: $D = \{x^{(i)}, y^{(i)}\}_{i=1}^{|D|}$, where each $x^{(i)}$ is a sequence of inputs, and each $y^{(i)}$ is a sequence of desired predictions. The estimation of weights of feature function Ө is typically performed by maximizing the conditional log-likelihood of annotated sequences of D as follows.

$$L(\Theta) = \sum_{i=1}^{|D|} \log P(y^i | x^i) \quad (3)$$

Despite the computational expense during CRF training [43], their main advantage lies in their power to represent much more knowledge than HMM thanks to feature functions. Thus, it is worth noting that, a HMM can be seen as a particular case of a CRF since the graph of the CRF model is built by the same links and nodes as the graph of the HMM. For an excellent tutorial covering more details about CRF, reader can see [11, 43].

For both approaches, the quality of used data -during the generation of sequences of observations- is a key factor for the success of CRF and HMM models.

## 5. WHY OBSERVATIONS ARE UNCERTAIN DATA

HMM and CRF are a classification PGM which make use of transition likelihoods between observable and hidden states to output a label for the current data point. It's really important that the quality of used sequences is crucial for the success of such models. Although, sequences of data used by both models reflect a real world process, this does not mean that these sequences have a perfect quality [44]. In fact, Cappé et al. [45] described the data used by such models as incomplete and inaccurate data and the observations resulting from these data are simply a subset of complete data. According to the authors, this poor quality of observation sequences is due to the uncertainty of used data and can be explained by the following two causes:

1) The effect of modeling physical process by a mathematical model. Indeed, passing from a real world process to a mathematical model is done through "imperfect" human interpretation which can cause imperfections in the model design.

2) Stochastic uncertainties due to variability of environment and measurement conditions [19]. HMM and CRF models operate using observation sequences.

Remember that each sequence of observations noted by X={$x_1$, $x_2$, ..., $x_T$} is a temporal stochastic sequence representing what is observed during a real world process. This sequence is built by integrating (at each time step t) just one symbol - of the alphabet $M = \{x_1, ..., x_N\}$ - which is the most representative of the process state (at this time step t). The conformity of the sequence of observations with the "true" real world process states is extremely important for the success of HMM and CRF models. In order to be more accurate in the construction of the observation sequences, it's really important that the observer (human or machine) makes the right choice of each observation symbol in order to describe "perfectly" the corresponding state of the process at time t, but the observer' choice depends on observer's knowledge often imperfect and/or incomplete. Usually, we assume that the observer has a perfect knowledge and a symbol will be instantiated- in order to be inserted into the sequence of observations - if a process state is determined with certainty by the observer. This strict correspondence between process state and observation symbol does not take into account the margin of error that can make the observer in his choice as described in section 1 by the example of athlete activity.

Using the membership relativity principle on which is based the fuzzy logic and possibility theory, it would be more appropriate to assign a degree of possibility and necessity for each process state in order to be represented by an observation symbol in the sequence of observations.

Although, improvements of data quality used by probabilistic graphical models significantly improve their performance, the consideration of uncertainty in order to deal with the "imperfect" quality of HMM and CRF used data remains a little discussed problem. According to the literature, the HMM improvement based on fuzzy logic [46-48] or belief functions [49] or the theory of evidence [50] have not taken into consideration the quality of used data during generation of sequences of observation. As well, a little effort was devoted to CRF improving using fuzzy logic [51, 52]. To the best of our knowledge, there is no previous work that has taken advantage of fuzzy logic and possibilistic theory in order to improve the quality of observation sequences for HMM and CRF models.

## 6. POSSIBILITY THEORY : A BRIEF PRESENTATION

*6.1 A brief historical overview*

The origins of the modalities "possible" and "necessary" date back to the middle ages [53] on Aristotle's and Theophrastus' works [54]. At the beginning of 20th century, these notions became the building blocks of modal logics [55] when possibility and necessity are all-or-nothing notions. More recently, between 40's and 70's, the economist G.L.S. Shackle [56] called the degree of potential surprise of an event its degree of impossibility which is valued on a disbelief scale. The impossibility of an event (necessity of the opposite event) in Shackle's view is understood as disbelief. A few years later, the philosopher David Lewis [57] considered a graded notion of possibility and introduced the comparative possibility in order to compare possible worlds. Then later the philosopher L.J.Cohen [58] considered the problem of legal reasoning by introducing the degree of provability known as the necessity measure. Based on the degree of provability of Cohen a hypothesis and its negation cannot both be provable together. Consequently, L. Zadeh was not the first to speak about formalizing notions of possibility but he was one of the most famous scientists of the 20th century who dealt with the uncertainty characterizing imprecise data by introducing the basic concepts of fuzzy sets and fuzzy logic in late 60's [20]. Although, Zadeh coined the term "Theory of Possibility", in his paper [21] in 1978 -to better represent and manipulate the "imperfect" quality of data- the term "possibilistic" was first introduced in 1975 by Gaines et al. [22]. In their paper Gaines et al. proposed the main foundations of the "logic of possibility", by the introduction of possibility, uncertainty and necessity concepts.

Based on "logic of possibility", Zadeh deals with fuzzy sets as a basis of theory of possibility [21]. Thereafter, Dubois and Prade [23, 59], further contributed to the development of the possibility theory and have widely used it to deal with uncertainty covering the incomplete knowledge. Although, fuzzy logic deals with imprecise data, possibility theory deals with uncertain knowledge. Since the imprecision and uncertainty are well linked, fuzzy sets theory and possibility theory are closely related.

*6.2 Basic concepts*

The importance of the theory of possibility stems from the fact that much of the information on which human decisions are based is possibilistic rather than probabilistic in nature [21]. In order to model the uncertainty of incomplete knowledge, the possibility distribution $\pi$ constitutes one of the fundamental concepts in possibilistic theory. Given a universe of discourse $\Omega = \{x_1, x_2, ..., x_n\}$, $\pi$ is a function which associates to each element $x_i$ from the universe of discourse $\Omega$ a value called possibility degree or degree of feasibility. This degree encodes our knowledge on the real world [60]. In possibility theory, the scale can be quantitative from the interval [0, 1] or qualitative in the form of an ordering between the different possible values that is important. For more details on discussions between qualitative and quantitative possibility theory, reader can see [53, 61]. In the present paper we deal with a quantitative possibility theory, thus the uncertainty scale used will be represented by the unit interval [0, 1].

$\pi(x_i)$ evaluates the plausibility that $x_i$ is the actual value of some variable to which $\pi$ is attached. $\pi(x_i) = 0$ means that $x_i$ is impossible ($x_i$ cannot be the real world) and $\pi(x_i) = 1$ means that $x_i$ is fully possible ($x_i$ is the real world). On possibility theory the two extreme cases of knowledge are given by:

1) The total ignorance : $\forall x_i \in \Omega, \pi(x_i) = 1$
2) The complete knowlegede : $\exists x_i \in \Omega, \pi(x_i) = 1$ and $\forall x_i \neq x_j, \pi(x_j) = 0$

The possibility distribution should ensure the following properties:

$\forall x_i \in \Omega ; \sup \pi(x_i) = 1$

$\pi(\emptyset) = 0$ and $\pi(\Omega) = 1$.

From a possibility distribution, two dual measures can be derived: possibility and necessity measures taking values into [0, 1]. So, knowing that our knowledge is represented by $\pi$, the possibility of an event A, noted $\Pi(A)$ evaluates at which level the event A is consistent with our knowledge while the necessity of an event A, noted $N(A)$ evaluates at which level the event A is certainly implied by our knowledge. Formally, given a universe of discourse $\Omega = \{x_1, x_2,..., x_n\}$ and a possibility distribution $\pi$ on $\Omega$, for any event $A \subseteq 2^{\Omega}$,

$\Pi(A) = \max_{x \in A} \pi(x)$

$N(A) = \min_{x \notin A} (1 - \pi(x)) = 1 - \Pi(A^c)$, with $A^c$ is the contrary event of A.

Thus, If $\Pi(A) = 0$ then $N(A^c) = 1$, that is, if an event A is impossible then its contrary is absolutely certain. $N(A) = 1 - \Pi(Ac)$ expresses the duality between possibility and necessity values. In possibility theory, the following rules must be satisfied:

$N(A) \leq \Pi(A)$
$N(A) > 0 \rightarrow \Pi(A) = 1$
Max $(\Pi(A), \Pi(A^c)) = 1$ ; one among A or $A^c$ is possible.
$\Pi(A \cup B) = \max (\Pi(A), \Pi(B))$ and $\Pi(A \cap B) \leq \min(\Pi(A), \Pi(B))$
$N(A \cap B) = \min (N(A), N(B))$ and $N(A \cup B) \geq \max (N(A), N(B))$

Possibility theory constitutes a powerful and simple alternative to probability theory in particular for dealing with uncertainty knowledge which is difficult to be represented by the probability theory. Therefore, the possibility theory has been widely used -especially during the two last decades- in order to simulate human reasoning in problem solving when information is uncertain. Various studies confirmed the high efficiency of the possibility theory in knowledge uncertainty representation such as in information retrieval process [62, 63], activity recognition of patients in a smart home [64] and many other studies [64, 65] where the possibility theory was used to recognize the goal of the users. Considering data uncertainty, possibilistic approach seems to be very efficient for problem solving where knowledge or measurements are "imperfect" such as sequences of observation generation.

## 7. THE PROPOSED MODEL

The fundamental basis of the proposed model is to deal with uncertainty during the instantiation of symbols representing "perfectly" the states of a given real world process. Thus, for representing a given state we rely on possibilistic quantification in order to choose "with certainty" the most representative symbol describing this observed state. The following is a detailed description of the proposed model.

Let $AOIs = \{A_1, ..., A_N\}$ be a set of N defined Areas Of Interest in the web interface and $F = \{f_1, f_2, ..., f_T\}$ is a finite sequence of recorded mouse fixations coordinates during a user task. Our approach is based on the assumption that each mouse cursor fixation $f_t$ has a fuzzy membership degree to each Area Of Interest $A_i$ in the used interface and each area $A_i$ can be represented as a fuzzy spatial entity [66] made up of a kernel called $ker(A_i)$, a region very close to the area's Kernel noted $Near(A_i)$ and a distant region from the kernel called $Far(A_i)$ as described in figure 2.

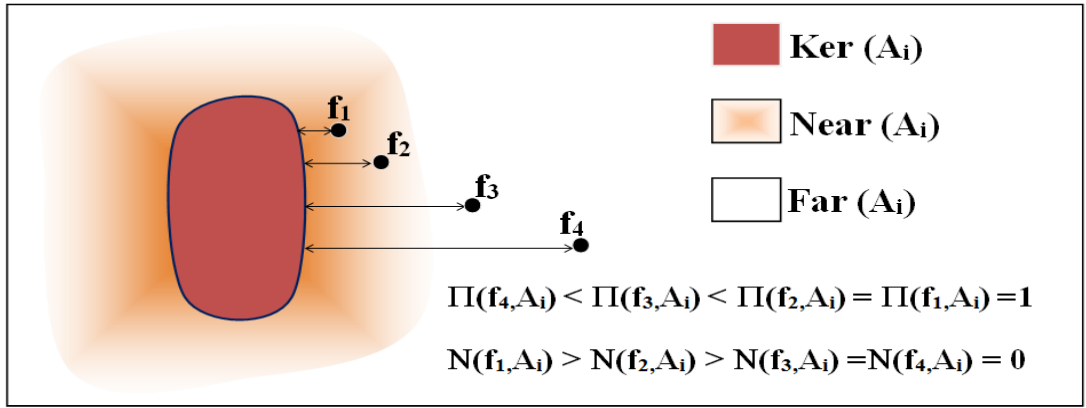

**Fig. 2. Possibility and nececity values increases if the fixation approaches the Ker(Ai)**

To evaluate the "perfect" belonging of a fixation $f_t$ (real world state) to an area $A_i$ (observation) we use a membership function $\mu_{Ai}(f_t)$ defined by:

$$\mu_{Ai}(f_t) = \begin{cases} \mu_{Ai}(f_t) = 1 & \text{if } f_t \in \text{Ker}(A_i) \\ 0 < \mu_{Ai}(f_t) < 1 & \text{if } f_t \in \text{Near}(A_i) \\ \mu_{Ai}(f_t) = 0 & \text{if } f_t \in \text{Far}(A_i) \end{cases} \quad (4)$$

In the classical approach [34, 40] each AOI ($A_i$) is made up of two parts which are Ker($A_i$) and Far($A_i$). Consequently, a fixation $f_t$ may either "belong to" the area $A_i$ or "not belong to" the area $A_i$. Using the function $\mu_{Ai}(f_t)$, we ensure that each fixation $f_t$ outside the kernel of an area $A_i$ -but close to it- may have a non-zero belonging degree to this area. To quantify the belonging degree of a fixation $f_t$ to an area $A_i$, we use possibility and necessity measures denoted respectively by $\Pi(f_t, A_i)$ and $N(f_t, A_i)$ as described in the following table.

**Table 1**
**Possibility and Necessity values according to fixation's position in different region of a fuzzy AOI.**

|  | The Cursor Fixation ($f_t$) is | | | | | |
|---|---|---|---|---|---|---|
|  | IN ($\in$) | | | NOT IN ($\notin$) | | |
|  | Ker($A_i$) | Near($A_i$) | Far($A_i$) | Ker($A_i$) | Near($A_i$) | Far($A_i$) |
| $\Pi(f_t, A_i)$ | =1 | =1 | $\in\,]0, 1[$ | $\in\,]0, 1]$ | $\in\,]0, 1]$ | =1 |
| $N(f_t, A_i)$ | =1 | $\in\,]0, 1[$ | =0 | $\in\,[0, 1[$ | =1 or =0 | $\in\,]0, 1]$ |

As shown in the table 1, each fixation $f_t \in \text{ker}(A_i)$ has a possibility value $\Pi(f_t, A_i) = 1$ and a necessity value $N(f_t, A_i) = 1$, that is, we are sure that the fixation $f_t$ certainly belongs to $A_i$. But if the fixation is not in the kernel ($f_t \notin \text{ker}(A_i)$) thus, the fixation $f_t$ is either very close to the kernel of the area ($A_i$) or it is far from it. In our proposed approach, these two cases are treated differently.

For the first case, where $f_t \notin \text{ker}(A_i)$ but $f_t \in \text{Near}(A_i)$, we assume that it is "fully" possible that the fixation $f_t$ belongs to $A_i$, but this is not a certain information. Thus the membership is evaluated by $\Pi(f_t, A_i)=1$ and $N(f_t, A_i) \in\,]0,1[$. The necessity value in this case increases if the fixation $f_t$ approaches the kernel Ker($A_i$) and decreases if the fixation $f_t$ gets away from Ker($A_i$) as described in figure 2.

For the second case, if $f_t$ is outside the kernel and outside the near region of the area $A_i$ ($f_t \notin \text{ker}(A_i)$ and $f_t \notin \text{Near}(A_i)$ and $f_t \in \text{Far}(A_i)$), we assume that $N(f_t, A_i) = 0$ and $\Pi(f_t, A_i) \in\,]0, 1[$ and the possibility value increases if the fixation approaches Near($A_i$) and decreases if the fixation $f_t$ moves away from Near($A_i$) as shown in figure 2. Based on this approach, the belonging of a given fixation $f_t$ to each area $A_i$ in the interface is weighted by necessity and possibility degrees as represented in figure 3.

In order to define the possibility $\Pi(f_t, A_i)$ and necessity $N(f_t, A_i)$ values of fixation $f_t$ belonging to a given area $(A_i)$, we assume that:

- $A_i$ : The Area Of Interest number i in the interface.
- $f_t$ : The mouse fixation at time t. with $1 \leq t \leq T$ and T is the total duration of the user task.
- $d(f_t, A_i)$ : The Euclidean distance between a fixation $f_t$ and the Kernel of area $A_i$.

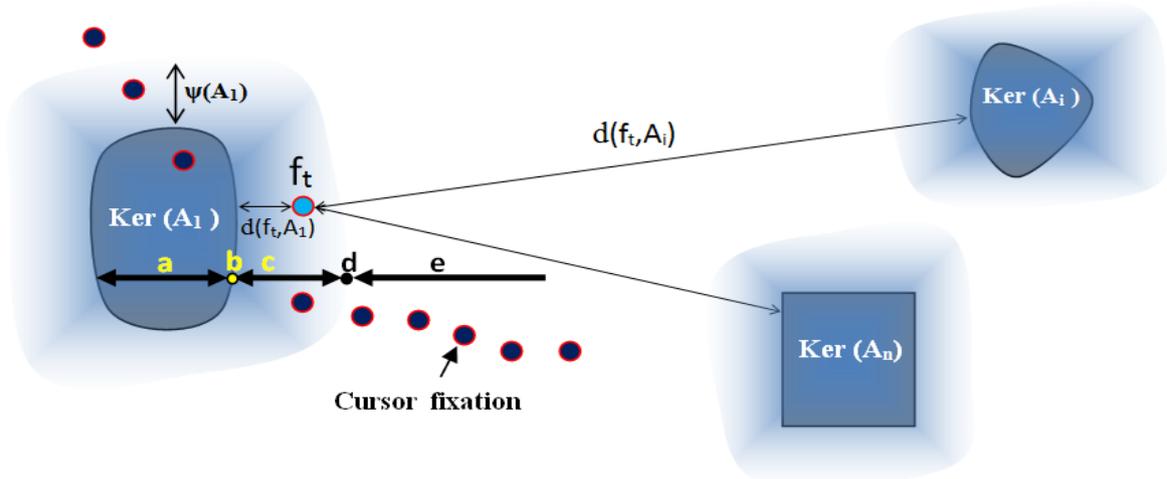

a) and b) $\Pi(f_t, A_1) = 1$ and $N(f_t, A_1) = 1$    c) $\Pi(f_t, A_1) = 1$ and $0 < N(f_t, A_1) < 1$
d) $\Pi(f_t, A_1) = 1$ and $N(f_t, A_1) = 0$    e) $0 < \Pi(f_t, A_1) < 1$ and $N(f_t, A_1) = 0$

Fig. 3. Cursor fixation membership possibility and necessity degrees

As described above in section 3, a particular region of the screen - such as a link, a textbox, an image or any other item in the interface - is defined as an Area Of Interest if it is frequently fixed (used) by users during their web activities. Consequently, each area of interest $A_i$ has an "attractive power" which determines to which degree this zone may attract the user's cursor. The attraction power value of each area $A_i$ is calculated based on the number of fixations recorded in this area. In fact, more fixations on a particular area indicate that this area is more noticeable or more important [39]. The attraction power of a given area $A_i$ is calculated as follows:

$$Atr_{Ai} = \frac{Total\ number\ of\ fixations\ in\ Ai}{Surface\ area\ of\ Ai} \quad (5)$$

Using (5), if the area $A_i$ is too small and it is frequently fixed during learner's activity, this shall indicate that this area draws the attention of the web user (and then mouse cursor) more than other areas having the same size but less fixed. To define the attraction power of a given area $A_i$ with respect to all other areas in the interface, we use the normalized attraction power $\lambda(A_i)$. The following formula (6) is defined for this purpose:

$$\lambda(A_i) = \frac{Atr_{Ai}}{\max_{k=1,2,...,N} Atr_{Ak}} \quad (6)$$

As described previously, $Near(A_i)$, is the closest region to the kernel of the area $A_i$. To define the proximity region $Near(A_i)$, we suppose that a higher attractive power $\lambda(A_i)$ has a positive effect on the size of the proximity region.

Therefore, an area with a higher attractive power, will have a Near($A_i$) region more important (wider) than another area with a lower attractive power. To define the distance $\psi(A_i)$ at which we consider a fixation as too close to the kernel of $A_i$ we use the following formula:

$$\psi(A_i) = \lambda(A_i) + \omega \qquad (7)$$

With $\omega$ is the distance (in pixel) to include in the vicinity region. The $\omega$ best value is defined by experimentation like shown below. So, using (7), attractive area has a larger "Near region" than a non-attractive area like described in the following figure.

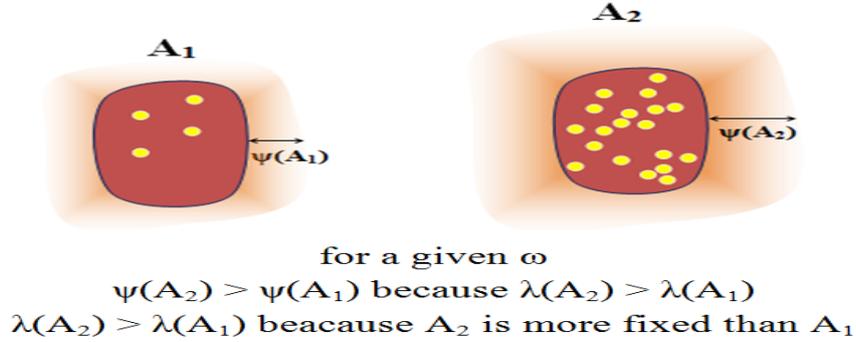

for a given $\omega$
$\psi(A_2) > \psi(A_1)$ because $\lambda(A_2) > \lambda(A_1)$
$\lambda(A_2) > \lambda(A_1)$ beacause $A_2$ is more fixed than $A_1$

**Fig. 4. The near region $\psi(A_i)$ distance according to the fixation frequency in Ker(Ai).**

In order to define the possibility and the necessity values to quantify the belonging degree of fixation $f_t$ to an area $A_i$, we suppose that each recorded fixation has a non-zero possibility value to belong in each area of the interface. Therefore, all recorded cursor fixations "possibly" belong to all areas of interest in the interface. Thus, if a fixation $f_t$ is in the ker($A_i$) or in the Near($A_i$) the possibility value $\Pi(f_t, A_i) = 1$. However, if a fixation $f_t$ is in Far($A_i$) region, the possibility value $\Pi(f_t, A_i) < 1$ and decreases if the fixation moves away from Near($A_i$) as shown if figure 2. In order to calculate $\Pi(f_t, A_i)$ in Far($A_i$) we use the following formula:

$$\Pi(f_t, A_i) = \frac{\Phi(f_t, A_i)}{d(f_t, A_i) / \psi(A_i)} \qquad (8)$$

With $\Phi(f_t, A_i)$ is the normalized "attachment degree" of fixation $f_t$ to an area $A_i$ calculated using (9).

Based on (8), closest to Near($A_i$) is the fixation $f_t$, higher is the possibility degree, and more distant -from Near($A_i$)- is the fixation, lower is the possibility degree $\Pi(f_t, A_i)$. In fact, if $d(f_t, A_i)$ increases (in other words if the fixation moves away from Near($A_i$)) then $\Pi(f_t, A_i)$ decreases and if $\psi(A_i)$ increases (because Near($A_i$) increases) then $\Pi(f_t, A_i)$ consequently increases. The attachment degree of a fixation to a given area $A_i$ indicates the degree to which a fixation is attracted by an area $A_i$. We normalize the attachment degree by dividing it by the max $U(f_t, A_k)$ in order to ensure that $\Pi(f_t, A_i)$ falls between 0 and 1.

$$\Phi(f_t, A_i) = \frac{U(f_t, A_i)}{\max_{k=1,2,...,N} U(f_t, A_k)} \qquad (9)$$

$\Phi(f_t, A_i)$ is calculated based on Bezdek formula [67].

$$\text{with} \quad U(f_t, A_i) = \frac{\left(\frac{1}{d(f_t, A_i)}\right)^{\frac{2}{m-1}}}{\sum_{k=1}^{N}\left(\frac{1}{d(f_t, A_k)}\right)^{\frac{2}{m-1}}} \ ; m>1 \qquad (10)$$

Based on Bezdek formula (10) the degree of belonging of a given fixation $f_t$ to the closest Kernel ker($A_i$), is significantly higher than the degree of belonging of the same fixation $f_t$ to the most distant Kernel ker($A_j$). Furthermore, the formula of Bezdek ensures that the degree of membership $U(f_t, A_k)$ is a normalized fuzzy value. Thus, using (10), closer is the fixation to a kernel of an area of interest, more attached to it is this fixation.

If a fixation is neither in the Ker($A_i$) nor in the Far($A_i$), consequently the fixation should be inside the Near($A_i$) region. For this case, $\Pi(f_t, A_i) = 1$ and we propose the following formula in order to calculate $N(f_t, A_i)$.

$$N(f_t, A_i) = \left(1 - \frac{d(f_t, A_i)}{\psi(A_i)}\right) * \frac{\min_{i=1,2,\ldots,N} d(f_t, A_i)}{d(f_t, A_i)} \quad (11)$$

Using (11), in the extreme case, when the fixation is located on the boundary separating the Near($A_i$) and Far($A_i$), le necessity value $N(f_t, A_i)$=0 (because $d(f_t, A_i) = \psi(A_i)$). This value increases if the fixation $f_t$ is within Near($A_i$) and approaches the ker($A_i$). however, if $f_t$ moves away from Ker($A_i$) to Far($A_i$) then $N(f_t, A_i)$ decreases.

Where the Near regions of two or more areas overlap and a fixation $f_t$ is located inside Near of two or more regions simultaneously, the formula (11) "attaches" the fixation $f_t$ to the area which has the higher attraction power. For better understanding this case, let us take the following example: let $A_1$ and $A_2$ be two AOI having respectively attraction powers 0,5 and 0,1. $A_1$ and $A_2$ are very close in such a way that their Near regions overlap as shown in the following figure.

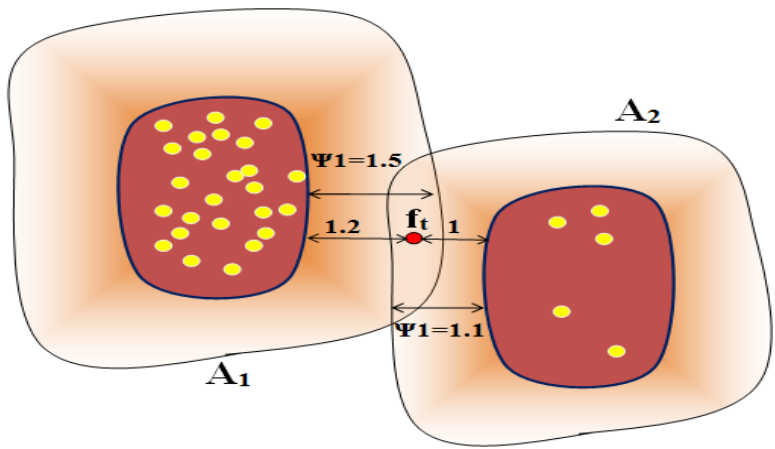

**Fig. 5 . Example of a fixation located in overlapped "Near" regions of two AOI .**

Our aim is to know which area ($A_1$ or $A_2$) will be instantiated (as an observation) in order to "better" represents the fixation $f_t$. The possibility values $\Pi(f_t, A_1) = 1$ and $\Pi(f_t, A_2) = 1$ does not reveal the relevant area that corresponds the best to the fixation $f_t$. Therefore, in order to solve this problem, we should calculate the necessities values $N(f_t, A_2)$ and $N(f_t, A_2)$.

For the example above, using formula (11) and for $\omega = 1$, we will get $N(f_t, A_1) = 0,16$ et $N(f_t, A_2) = 0,09$. Thus, although the fixation $f_t$ is closer to the kernel of $A_2$ (because $d(f_t, A_2)$=1) than the kernel of $A_1$ (because $d(f_t, A_1)$=1,2). Therefore, the fixation $f_t$ is much more "attached to" the area $A_1$ than $A_2$ because the attraction power of $A_1$ (0,5) is significantly higher than the attraction power of $A_2$ (0,1). That makes sense because most users have a tendency to fix the area $A_1$ more than area $A_2$ in order to perform the required task, consequently the fixation $f_t$ is more attracted by the Area $A_1$ than $A_2$.

In order to generate sequences of observations based on the proposed approach we use algorithm2 described below, while algorithm1 is used to generate classical observation sequences.

Algorithm 2: Possibilistic Vectorization Algorithm

---

**Vectorization algorithm2:**
// Possibility and necessity membership degree

*Begin vectorization*

*Initializations*
  $X^*=\{\}$ ;
  Details of each Area Of Interest ($A_i$);
  T ← Total duration of the task;
  ds ← Time between recordings of two cursor fixations;
  m ← Bezdek fuzzification value;
  ω ← distance of proximity (in pixel);
  P ← the lower possibility value to consider a fixation as relevant;

*Input:*
  Mouse trajectory {(x, y) coordinates of each fixation}.

*Treatment:*
  For each Area Of Interest $A_i$ do
    Calculate $\lambda(A_i)$ : normalized attractive power of $A_i$ using **(5)** and **(6)**
    Calculate $\psi(A_i)$ : proximity of $A_i$ using **(7)**
  End for

  For t:=1 to T (with ds step) do
    For each Area of interest $A_i$ do
      Calculate the Euclidean distance $d(f_t, A_i)$
      If ( fixation $f_t \in ker(A_i)$ ) then
        Begin
          **$N(f_t, A_i)$ ← 1;**
          **$\Pi(f_t, A_i)$ ← 1;**
        End

      ElseIf ( fixation $f_t \in Near(A_i)$) then
       Begin
        **Calculate $N(f_t, A_i)$ using (11)**
       **$\Pi(f_t, A_i)$ ← 1**
       End

      ElseIf ( fixation $f_t \in Far(A_i)$) then
       Begin
        **$N(f_t, A_i)$ ← 0**
        **Calculate $\Pi(f_t, A_i)$ using (8)**
       End
      End if
    End For

    For each Area of Interest $A_i$ do
      If $0 < N(f_t, A_i) \leq 1$ then
        $X^*[t]=argmax(N(f_t, A_i))$
        Else if $P < \Pi(f_t, A_i) \leq 1$
          $X^*[t]=argmax(\Pi(f_t, A_i))$
      End if
    End For

  End for

*Output:*
$X^* = \{x^*_1, x^*_2, ..., x^*_T\}$ //**possibilistic observation sequence**

*End vectorization*

Algorithm 1: Boolean Vectorization Algorithm

**Vectorization algorithm1**:
//Strict membership degree: 1 if ft in $A_i$ and 0 otherwise
*Begin vectorization*
*Initializations*
   X={ } ;
   Details of each kernel of Area Of Interest ($A_i$);
   T ← Total duration of the task;
   ds ← Time between recordings of two cursor fixations;
*Input:*
   Mouse trajectory {(x, y) coordinates of each fixation}.
*Treatment*:
For t:=1 to T (with ds step) do
    **If ( fixation ($f_t$) is in the kernel of Area ($A_i$)) then**
      **X[t] ← $A_i$**
    **End if**
End for
*Output:*
X = {$x_1$, $x_2$, ..., $x_T$}  // **classical observation sequence**

*End vectorization*

Figure 6 summarizes the task recognition process using classical and possibilistic sequences of observation.

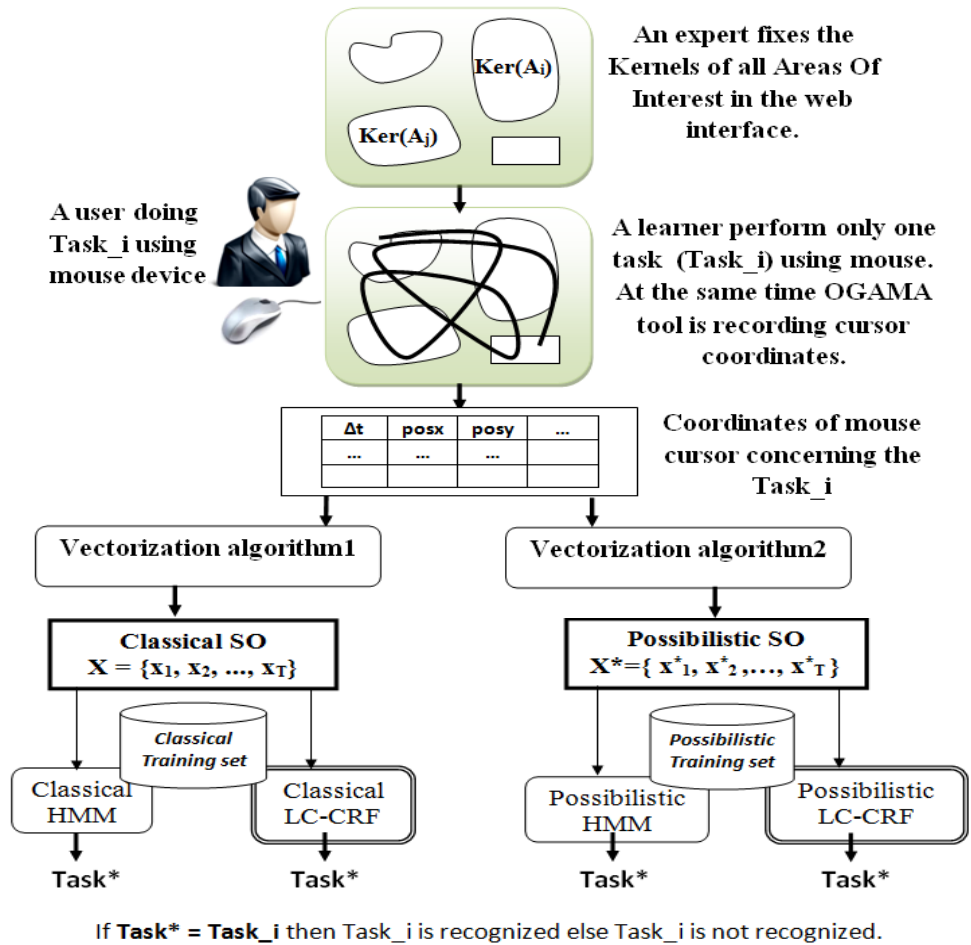

**Fig. 6 . User task recognition based on cursor movement data using classical and possibilistic sequences of observation.**

## 8. EXPERIMENTAL RESULTS AND DISCUSSION

### 8.1 Experimental setup

Our algorithms had been tested on recognizing three different tasks performed by 51 students using the "Equation Grapher" online simulator[1]. All students participated in experiments using the mouse device. Each student performs only one task among the three tasks detailed below.

The following figure represents the used e-learning interface as a set of 15 kernels of Areas Of Interest. An interface item is deemed by the expert as an important area if it is frequently fixed by mice of learners. Formally, the "Equation Grapher" interface can be described by AOIs={ ker(A), ker(B), … , ker(O)}.

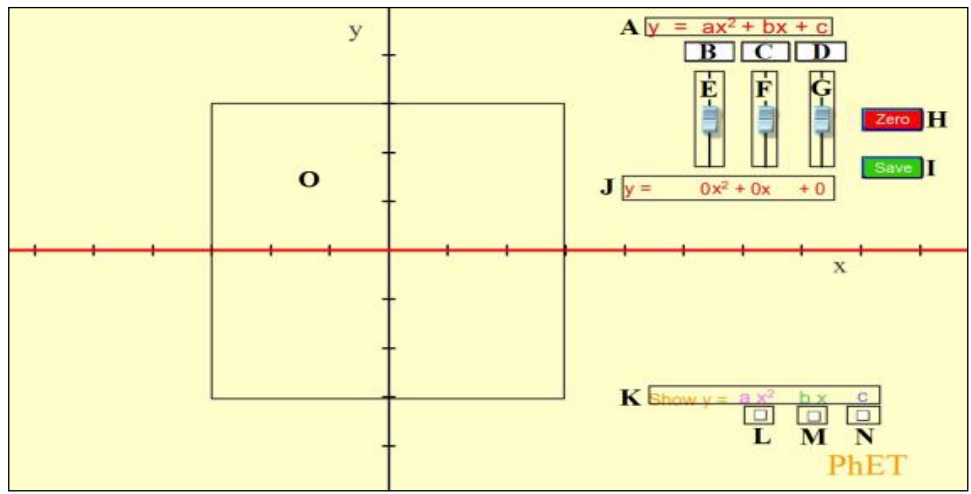

**Fig. 7. Kernels of Areas Of Interest in "Equation Grapher" interface.**

During task performing and without being aware, users move mouse cursor across the web interfaces and fix various items of interest. Thus, each task can be defined as a finite, temporal, stochastic sequence of areas of interest fixed during interaction process.

Before preparing any sequence of observations, we used OGAMA tool[2] for recording mouse cursor coordinates at each time slice (1 centisecond) during the user activity. The obtained sequence of data (vector of cursor coordinates) will be provided as input for algorithm1 in order to produce a Classical Sequence of Observation (CSO). The same sequence of recorded data will be provided for algorithm2 to produce a Possibilistic Sequence of Observation (PSO) like shown in figure 6.

During each task, the OGAMA tool is first setup in order to track and to record the mouse cursor movement, and then each student is asked to perform only one task. Each task type, is performed by a group of 17 learners. The three tasks are described as follows:

Task1(DEG2) : during this task a student is asked to perform a graphical representation of a quadratic equation of the form $ax^2+bx+c=0$ (with a, b and c $\neq 0$) and to keep in memory the shape of the drawn curve.

Task2 (DEG1): during this task a student is asked to perform a graphical representation of a quadratic equation of the form $ax^2+bx+c=0$ (with a=0 and b$\neq$0, c $\neq$0) and to keep in memory the shape of the drawn curve.

Task3(INT):  during this task, a student is asked to discover (and keep in memory) intersection coordinates of two quadratic equations of the form $ax^2+bx+c=0$. The first have as parameters a$\neq$0, b$\neq$0 and c$\neq$0, while the second have as parameters a=0 and c=0.

Once learners' mice movements were recorded, we prepare the sequences of observations for models parameters estimation and testing phases. Therefore the 51 recorded tasks are used to prepare two training sets. The first one included 51 possibilistic sequences of observations prepared using algorithm1, and the second containing 51 classical sequences of observations made using algorithm2. In order to define, to train and to test the HMM model we used the matlab toolbox[3] and for the CRF model description, training and inference we used the CRF++ tool[4].

---

[1] Phet available on : http://phet.colorado.edu

[2] OGAMA available on : http://www.ogama.net/

To validate the proposed model, we used the sampling technique "Leave One Out Cross Validation" (LOOCV) which has the advantage of using all sequences of observation for both training and validation. In fact, using LOOCV consisting of taking at each time one single task (represented by one sequence of observation) from the "training and test" set for testing and the remaining tasks are used for training.

The next section presents the obtained results and discusses the effect of possibilistic versus classical reasoning on HMM and CRF effectiveness in the e-learning task recognition based only on cursor movement tracking.

### 8.2 Experimental results

To evaluate HMM and CRF models performance we use the accuracy rate value defined as the ratio of the number of observation sequences which matches the ground truth to the total number of test sequences. Table 2 shows that, HMM recognition rate based on classical observation sequences (CHMM) is 76.47% and the recognition rate of the HMM based on possibilistic observation sequences (PHMM) is 88.23%.

**Table 2**
**HMM recognition rate using classical and possibilistic observation sequences.**

| Task Type | Classical observation sequences | | | Possibilistic observation sequences ($\omega=3$, $m=2$) | | |
|---|---|---|---|---|---|---|
| | Samples | Errors | CHMM accuracy rate | Samples | Errors | PHMM accuracy rate |
| Task1 (DEG2) | 17 | 8 | 52,94% | 17 | **4** | **76,47%** |
| Task2 (DEG1) | 17 | 0 | 100% | 17 | 0 | 100,00% |
| Task3 (INT) | 17 | 4 | 76,47% | 17 | **2** | **88,32%** |
| Total | 51 | 12 | **76,47%** | 51 | 6 | **88,23%** |

Similarly, in table 3 the accuracy rate of CRF model based on classical observation sequences (CCRF) is 88.23% whereas the CRF model recognition rate based on possibilistic observation sequences (PCRF) is 90.19%.

**Table 3**
**CRF recognition rate using classical and possibilistic observation sequences.**

| Task Type | Classical Observation Sequences | | | Possibilistic Observation Sequences ($\omega=8$, $m=2$) | | |
|---|---|---|---|---|---|---|
| | Samples | Errors | CCRF accuracy rate | Samples | Errors | PCRF accuracy rate |
| Task1 (DEG2) | 17 | 3 | 82,35% | 17 | 3 | 82,35% |
| Task2 (DEG1) | 17 | 2 | 88,23% | 17 | **1** | **94,11%** |
| Task3 (INT) | 17 | 1 | 94,11% | 17 | 1 | 94,11% |
| Total | 51 | 6 | **88,23%** | 51 | 5 | **90,19%** |

For PHMM and PCRF models, the best performance is reached after the determination of the ω value (distance in pixel of the vicinity extent) for each model. To find the best value ω, we regenerate a new set of possibilistic observation sequences for each ω value and we readjust PHMM and PCRF models parameters. Then we run the inference process. The following figure shows PHMM versus PCRF models accuracy rates according to ω values for each model.

---

[3] http://www.mathworks.com/help/stats/hidden-markov-models-hmm.html

[4] CRF++ available on: http://crfpp.googlecode.com/svn/trunk/doc/index.html

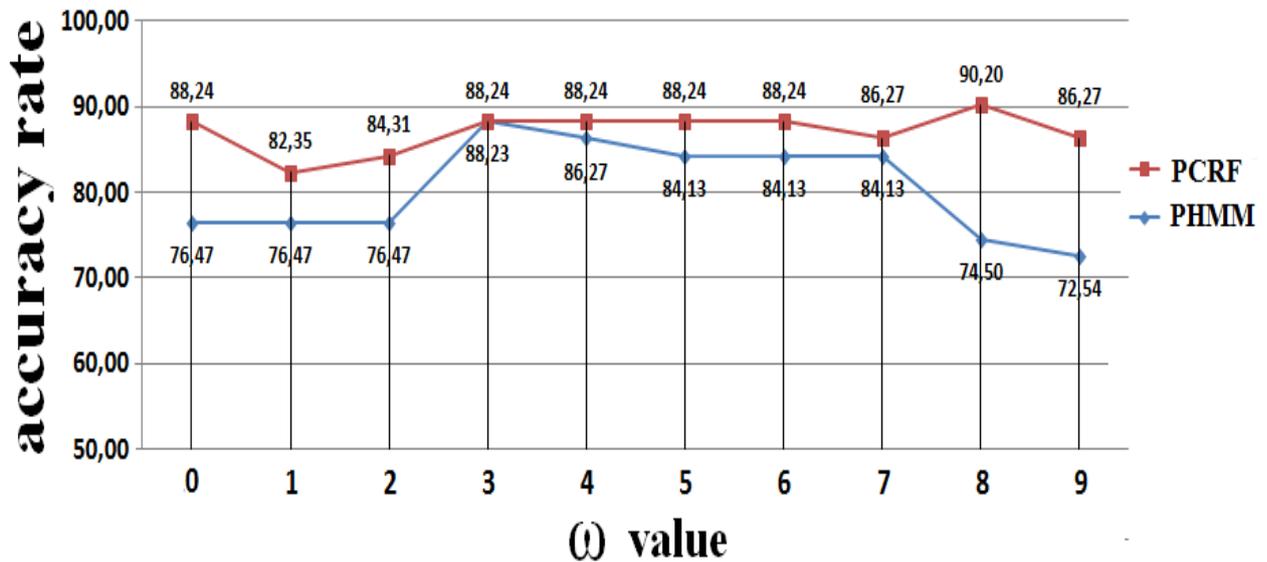

**Fig. 8. Possibilistic HMM and Possibilistic CRF accuracy rates according to ω value**

According to figure 8, when considering fixations which are in kernels of the areas of interest (ω = 0), the recognition rate of classical CRF (88,35%) is better than the recognition rate of classical HMM (76,47%), this confirms the findings of Elbahi et al.[35] in which authors confirm that CRF outperforms the HMM in user task recognition using a classical sequences of observations.

Likewise, if ω = 3 the recognition rate of PHMM (88,23%) is much better than classical HMM (76,47%). While for ω = 8 the recognition rate of PCRF (90,20%) is better than the classical CRF(88,24%).

According to the curves in figure 8, the HMM recognition capacity is more sensitive to the proposed approach than CRF model. In fact, depending on the ω value the PHMM recognition accuracy varies significantly more than PCRF recognition rate. This can be explained by the fact that HMM and CRF models operate in a very different way [68] despite some similarities between them. The PHMM best accuracy rate is reached when ω=3 and PCRF best accuracy rate is reached for ω=8. This is a result of the way both models maximize their parameters. The following section discusses the obtained results.

*8.3 Discussion*

The proposed model can be generalized to other types of web users' activities such as e-commerce and e-banking. Figure 8 shows the variation of the PHMM and PCRF recognition rates according to the value of ω. Indeed, if ω= 0 ($f_t \in ker(A_i)$), the fixations outside the kernels of areas of interest are totally ignored during sequences of observations preparation. For this case (ω= 0) the PHMM model keeps the same recognition rate (76.47%) than HMM model using classical observation sequences and PCRF model keeps the same recognition rate (88.24%) than CRF model using classical observation sequences.

The first finding is that the proposed approach retains the characteristics of classical models if ω =0. Hence, the classical approach is a particular case of proposed possibilistic approach. But when we deal with fixations which are very close to the kernel of areas of interest ($f_t \in Near(A_i)$), the recognition rate for both models changes significantly. So, if ω = 3 we find that the performance of PHMM model is maximum (88,23%) and if $4 \leq \omega \leq 7$ the recognition rate is better than classical HMM and for ω > 7 pixels, PHMM accuracy rate is lower than HMM using classical sequences. As well, PCRF model reached its maximum accuracy (90,20%) if ω=8 whereas for other values of ω the PCRF recognition rate is between 82,35% and 88,24%

The second finding is that possibilistic quantification of fixations belonging to different areas in the interface has a positive impact on probabilistic graphical models effectiveness especially HMM and CRF models. Indeed, as tables 2 and 3 make clear, when probabilistic models use observation sequences obtained based on possibilistic approach their accuracy will be improved compared with the same models using classical observation sequences.

Another important remark concerning the recognition of task2. For this DEG1 task, classical HMM performs better (100%) than classical CRF (88,23%), but thanks to possibilistic reasoning (during sequences of observation preparation) the PCRF becomes able to recognize one other DEG1 task achieving a higher accuracy rate (90,20%) than classical CRF model.

To explain why HMM and CRF behave differently according to $\omega$ value, it should be recalled that, despite the similarities between both models, they operate differently during training and inference process. In fact, HMM are generative models in which an observation is supposed to be independent of all other variables, given the hidden variable. Moreover, HMM make use of bayesian framework in which a separate model is learned for each task and the posterior probability can be calculated for each novel cursor fixation. While CRF are discriminative models which relax the conditional independence assumption of HMM and just one model is used by CRF model for all tasks, by calculating $p(y|x)$ directly. As already introduced, all parameters of CRF are re-estimated by maximizing the conditional likelihood $p(y|x)$ instead of the joint probability $P(Y, X)$ as do HMM. Therefore, due to the single model used by CRF for all tasks, tasks compete during maximization. Therefore, using a dataset with different tasks (that use same areas of interest to achieve quite similar goals), modeling everything as the dominant task might yield a higher likelihood than including and/or misclassifying parts of some tasks. This view is confirmed by the identification capability of task2 by classical and possibilistic models. Indeed, the HMM model is more effective (100%) than CRF model (88,23%) and the PHMM model (100%) is more effective than PCRF (94,11%).

## 9. CONCLUSION AND FUTURE WORK

For years, HMM and CRF approaches had gained significant popularity thanks to their good performance in classification tasks and stochastic sequences labeling. To be useful in real world applications, such models depend mainly on the quality or real data during sequences of observation generation. Despite imperfect quality of used data, HMM and CRF models are characterized by their good efficiency and their simple applicability.

In this paper we proposed to consider uncertainty of data during observation sequences construction. Each sequence of observation is related to a specific user experiment, in which students are asked to interact with an online simulator using a mouse to perform a given task.

Although the effectiveness of the proposed approach in the preparation of data sequences for HMM and CRF models, experimental results showed that possibilistic approach reasoning had a significant improvement in HMM and CRF efficiency. In addition to the enhancement of the effectiveness of HMM and CRF models based on possibilistic reasoning, the proposed approach maintained the superiority of the CRF approach compared to the HMM in e-learning tasks recognition using mouse movement tracking technique. The new approach is a general technique and it can be easily applied to other online activities. However, despite its apparent simplicity and its proven effectiveness, it is clear that the determination of the optimal value of the proximity ($\omega$) for each HMM and CRF models remains the main limitation of the proposed approach due to the time consuming constraint.

As future works, to overcome this shortcoming we plan to use an optimization algorithm in order to determine more quickly and automatically the optimal value of $\omega$ using much larger datasets.